# Iris Recognition Based on SIFT Features

Fernando Alonso-Fernandez, Pedro Tome-Gonzalez, Virginia Ruiz-Albacete, Javier Ortega-Garcia

*Abstract—* Biometric methods based on iris images are believed to allow very high accuracy, and there has been an explosion of interest in iris biometrics in recent years. In this paper, we use the Scale Invariant Feature Transformation (SIFT) for recognition using iris images. Contrarily to traditional iris recognition systems, the SIFT approach does not rely on the transformation of the iris pattern to polar coordinates or on highly accurate segmentation, allowing less constrained image acquisition conditions. We extract characteristic SIFT feature points in scale space and perform matching based on the texture information around the feature points using the SIFT operator. Experiments are done using the BioSec multimodal database, which includes 3,200 iris images from 200 individuals acquired in two different sessions. We contribute with the analysis of the influence of different SIFT parameters on the recognition performance. We also show the complementarity between the SIFT approach and a popular matching approach based on transformation to polar coordinates and Log-Gabor wavelets. The combination of the two approaches achieves significantly better performance than either of the individual schemes, with a performance improvement of 24% in the Equal Error Rate.

## I. INTRODUCTION

Recognizing people based on anatomical (e.g., fingerprint, face, iris, hand geometry, ear, palmprint) or behavioral characteristics (e.g., signature, gait, keystroke dynamics), is the main objective of biometric recognition techniques [1]. The increasing interest on biometrics is related to the number of important applications where a correct assessment of identity is a crucial point. Biometric systems have several advantages over traditional security methods based on something that you know (password, PIN) or something that you have (card, key, etc.). In biometric systems, users do not need to remember passwords or PINs (which can be forgotten) or to carry cards or keys (which can be stolen). Among all biometric techniques, iris recognition has been traditionally regarded as one of the most reliable and accurate biometric identification system available [2]. Additionally, the iris is highly stable over a person's lifetime and lends itself to noninvasive identification because it is an externally visible internal organ [3].

Traditional iris recognition approaches approximates iris boundaries as circles. The ring-shaped region of the iris is then transferred to a rectangular image in polar coordinates as shown in Figure 1, with the pupil center being the center of the polar coordinates [4]. This transfer normalizes the distance between the iris boundaries due to contraction/dilation of the pupil, the camera zoom or the camera

Biometric Recognition Group - ATVS, Escuela Politecnica Superior, Universidad Autonoma de Madrid, Avda. Francisco Tomas y Valiente, 11, Campus de Cantoblanco, 28049 Madrid, Spain, email: {fernando.alonso, pedro.tome, virginia.ruiz, javier.ortega}@uam.es

to eye distance. When converting an iris region to polar coordinates, it is necessary a very accurate segmentation in order to create a similar iris pattern mapping between images of the same eye [5]. Features are then extracted from the rectangular normalized iris pattern. For this purpose, a number of approaches have been proposed in the literature [6], e.g.: Gabor filters, log-Gabor filters, Gaussian filters, Laplacian-of-Gaussian filters, wavelet transforms, etc.

One of the drawbacks of traditional iris recognition approaches is that the transformation to polar coordinates can fail with non-cooperative or low quality data (e.g. changes in the eye gaze, non-uniform illumination, eyelashes/eyelids occlusion, etc.) [5]. In this paper, we implement the Scale Invariant Feature Transformation (SIFT) [7] for its use in biometric recognition using iris images. SIFT extracts repeatable characteristic feature points from an image and generates descriptors describing the texture around the feature points. The SIFT technique has already demonstrated its efficacy in other generic object recognition problems, and it has been recently proposed for its use in biometric recognition systems based on face [8], [9], fingerprint [10] and iris images [5]. One of the advantages of the SIFT approach is that it does not need transfer to polar coordinates. We have used for our experiments the BioSec multimodal baseline corpus [11] which includes 3,200 iris images from 200 individuals acquired in two different sessions. We analyze the influence of different SIFT parameters on the verification performance, including the implementation of a technique to remove false matches, as proposed previously for fingerprints [10]. We also demonstrate that the proposed approach complements traditional iris recognition approaches based on transformation to polar coordinates and Log-Gabor wavelets [12], [13]. In our experiments, the fusion of the two techniques achieves a performance improvement of 24% in the Equal Error Rate.

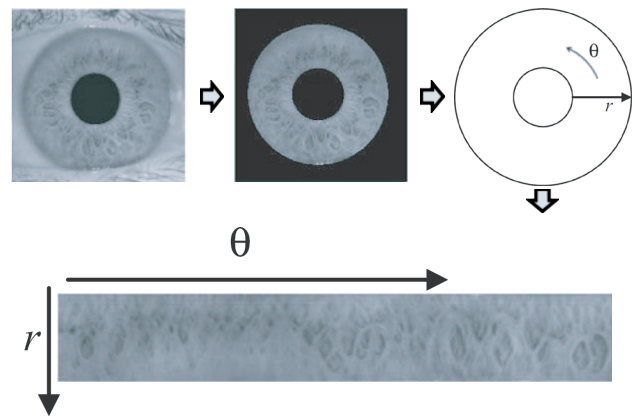

Fig. 1. Normalization of the iris region to polar coordinates. The ring-shaped region of the iris is transferred to a rectangular image, with the pupil center being the center of the polar coordinates.

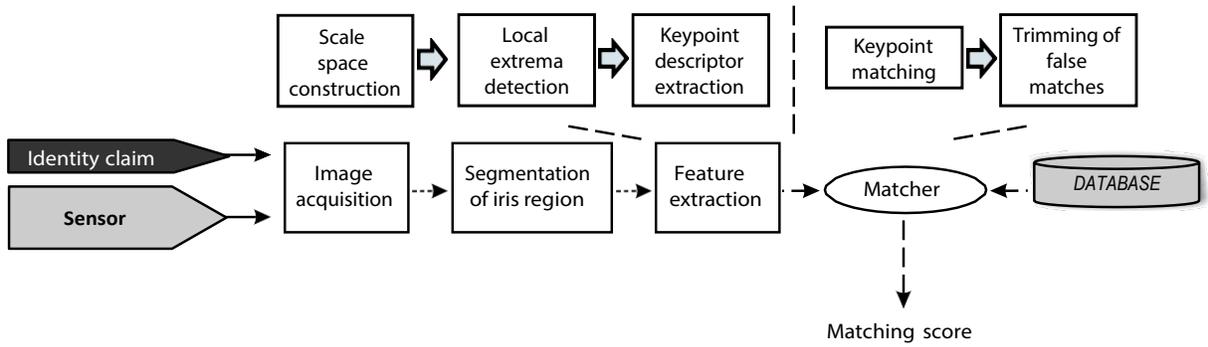

Fig. 2. Architecture of an automated iris verification system using the SIFT operator.

Furthermore, since the SIFT technique does not require polar transformation or highly accurate segmentation, and it is invariant to changes in illumination, scale and rotation, it is hoped that this technique will be feasible with unconstrained image acquisition conditions. One of the major current practical limitations of iris biometrics is the degree of cooperation required on the part of the person whose image is to be acquired. All existing commercial iris biometrics systems still have constrained image acquisition conditions [6]. Current efforts are aimed at acquiring images in a more flexible manner and/or being able to use images of more widely varying quality, e.g. the "Iris on the Move" project [14], which is aimed to acquire iris images as a person walks at normal speed through an access control point such as those common at airports. This kind of systems would drastically reduce the need of user's cooperation, achieving transparent and low-intrusive biometric systems, with a higher degree of acceptance among users.

The rest of the paper is organized as follows. Section II describes the SIFT algorithm. Section III describes our experimental framework, including the database used, the protocol, and the results. Finally, conclusions and future work are drawn in Section IV.

## II. SCALE INVARIANT FEATURE TRANSFORMATION (SIFT)

Scale Invariant Feature Transformation (SIFT) [7] was originally developed for general purpose object recognition. SIFT detects stable feature points of an object such that the same object can be recognized with invariance to illumination, scale, rotation and affine transformations. A brief description of the steps of the SIFT operator and their use in iris recognition is given next. The diagram of a iris recognition system using the SIFT operator is shown in Figure 2.

### A. Scale-space local extrema detection

The first step is to construct a Gaussian scale space, which is done by convolving a variable scale 2D Gaussian operator $G(x, y, \sigma)$ with the input image $I(x, y)$:

$$L(x, y, \sigma) = G(x, y, \sigma) * I(x, y) \quad (1)$$

Difference of Gaussian (DoG) images $D(x, y, \sigma)$ are then obtained by subtracting subsequent scales in each octave:

$$D(x, y, \sigma) = L(x, y, k\sigma) - L(x, y, \sigma) \quad (2)$$

where $k$ is a constant multiplicative factor in scale space. The set of Gaussian-smoothed images and DoG images are called an octave. A set of such octaves is constructed by successively down sampling the original image. Each octave (i.e., doubling of $\sigma$) is divided into an integer number $s$ of scales, so $k = 2^{1/s}$. We must produce $s+3$ images for each octave, so that the final extrema detection covers a complete octave. In this paper we have used $s=3$, thus producing six Gaussian-smoothed images and five DOG images per octave, and a value of $\sigma=1.6$ (values from Lowe [7]). Figure 3 shows 3 successive octaves with 6 scales and the corresponding difference images.

Local extrema are then detected by observing each image point in $D(x, y, \sigma)$. A point is decided as a local minimum or maximum when its value is smaller or larger than all its surrounding neighboring points. Each sample point in $D(x, y, \sigma)$ is compared to its eight neighbors in the current image and nine neighbors in the scale above and below.

### B. Accurate Keypoint Localization

Once a keypoint candidate has been found, if it observed to have low contrast (and is therefore sensitive to noise) or if it is poorly localized along an edge, it is removed because it can not be reliably detected again with small variation of viewpoint or lighting changes. Two thresholds are used, one to exclude low contrast points and other to exclude edge points. More detailed description of this process can be found in the original paper by Lowe [7].

### C. Orientation assignment

An orientation histogram is formed from the gradient orientations within a 16x16 region around each keypoint. The orientation histogram has 36 bins covering the 360 degree range of orientations. Each sample added to the histogram

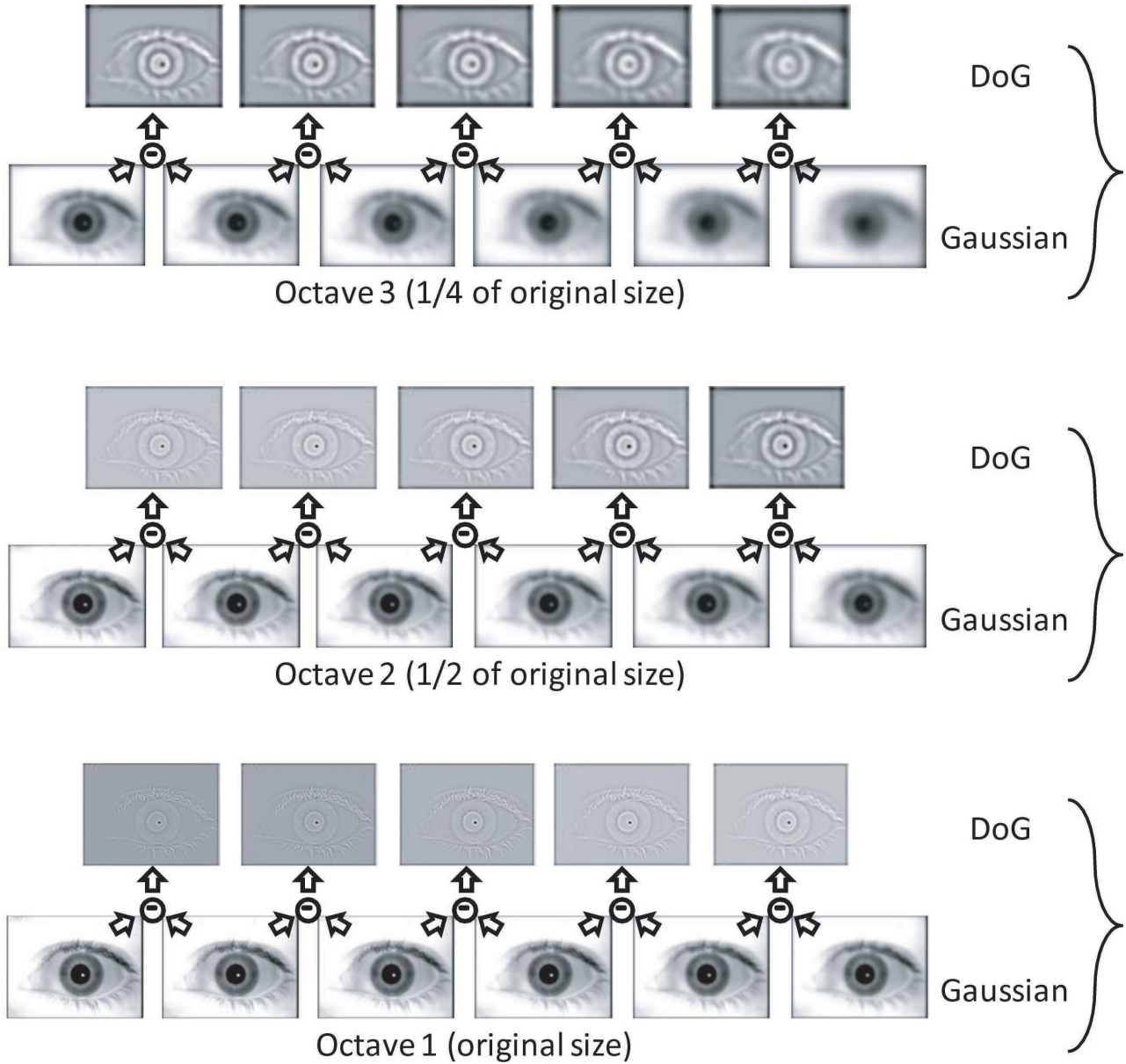
Fig. 3. Example of SIFT scale space construction. The figure shows 3 successive octaves, with 6 scales per octave, and the corresponding difference images.

The highest peak in the histogram is then detected, as well as any other local peak that is within 80% of the highest peak. For locations with multiple peaks, there will be multiple keypoints created at the same location, but with different orientations. The major orientations of the histogram are then assigned to the keypoint, so the keypoint descriptor can be represented relative to them, thus achieving invariance to image rotation.

### D. Keypoint descriptor

In this stage, a distinctive descriptor is computed at each keypoint. The image gradient magnitudes and orientations, relative to the major orientation of the keypoint, are sampled within a 16×16 region around each keypoint. These samples are then accumulated into orientation histograms summarizing the contents over 4×4 subregions, as shown in Figure 4. Each orientation histogram has 8 bins covering the 360 degree range of orientations. Each sample added to the histogram is weighted by its gradient magnitude and by a Gaussian circular window centered at the local extremum. The descriptor is then formed from a vector containing the values of all the orientation histogram entries,

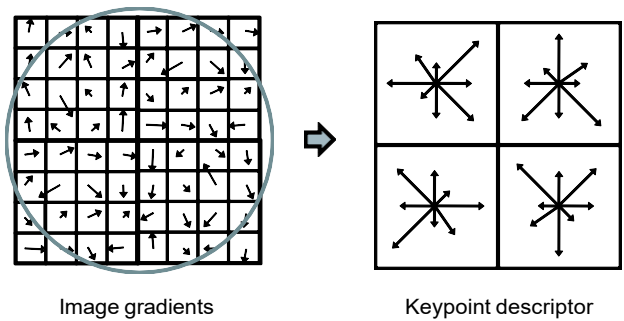

Fig. 4. Computation of SIFT keypoint descriptor (image from [7]). The gradient magnitude and orientation at each image sample point in a region around the keypoint location is first computed, as shown on the left, weighted by a Gaussian window (indicated by the overlaid circle). These samples are then accumulated into orientation histograms summarizing the contents over 4×4 subregions, as shown on the right, with the length of each arrow corresponding to the sum of the gradient magnitudes near that direction within the region. The figure shows a 2×2 descriptor array computed from an 8×8 set of samples, whereas the experiments in this paper use 4×4 descriptors computed from a 16×16 sample array.

therefore having a 4×4×8=128 element feature vector for each keypoint.

*E. Keypoint matching*

Matching between two images $I_1$ and $I_2$ is performed by comparing each local extrema based on the associated descriptors. Given a feature point $p_{11}$ in $I_1$, its closest point $p_{21}$, second closest point $p_{22}$, and their Euclidean distances $d_1$ and $d_2$ are calculated from feature points in $I_2$. If the ratio $d_1/d_2$ is sufficiently small, then points $p_{11}$ and $p_{21}$ are considered to match. Then, the matching score between two images can be decided based on the number of matched points. According to [7], we have chosen a threshold of 0.76 for the ratio $d_1/d_2$.

*F. Trimming of false matches*

The keypoint matching procedure described may generate some erroneous matching points. We have removed spurious matching points using geometric constraints [10]. We limit typical geometric variations to small rotations and displacements. Therefore, if we place two iris images side by side and draw matching lines as shown in Figure 5 (top), true matches must appear as parallel lines with similar lengths. According to this observation, we compute the predominant orientation $\theta_P$ and length $\ell_P$ of the matching, and keep the matching pairs whose orientation $\theta$ and length $\ell$ are within predefined tolerances $\varepsilon_\theta$ and $\varepsilon_\ell$, so that $|\theta - \theta_P| < \varepsilon_\theta$ and $|\ell - \ell_P| < \varepsilon_\ell$. The result of this procedure is shown in Figure 5 (bottom).

## III. EXPERIMENTAL FRAMEWORK

*A. Database and protocol*

For the experiments in this paper, we use the BioSec baseline database [11]. It consists of 200 individuals acquired in two acquisition sessions, separated typically by one to four weeks. A total of four iris images of each eye, changing eyes between consecutive acquisitions, are acquired in each session. The total number of iris images is therefore: 200 individuals $\times$ 2 sessions $\times$ 2 eyes $\times$ 4 iris = 3,200 iris images. We consider each eye as a different user, thus having 400 users. Glasses were removed for the acquisition, while the use of contact lenses was allowed. The database have been acquired with the LG Iris Access 3000 sensor, with an image size of 640 pixels width and 480 pixels height. Some iris examples are shown in Figure 6.

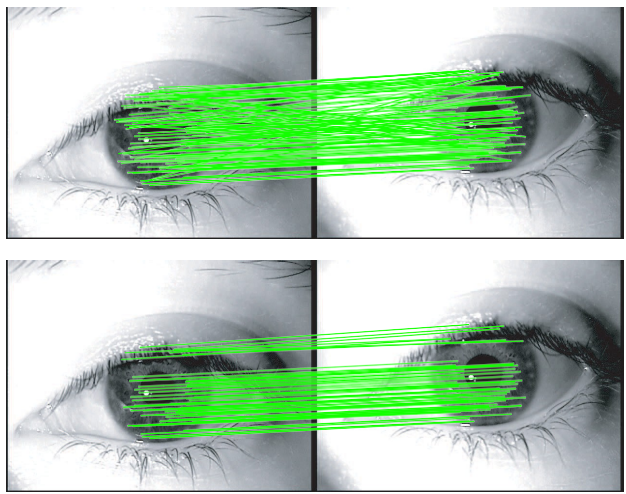

Fig. 5. Matching of two iris images using SIFT operators without and with trimming of false matches using geometrical constraints (top and bottom, respectively). Trimming of false matches is done by removing matching pairs whose orientation and length differ substantially from the predominant orientation and length computed from all the matching pairs.

The 200 subjects included in BioSec Baseline are further divided into [11]: *i*) the *development set*, including the first 25 and the last 25 individuals of the corpus, totaling 50 individuals; and *ii*) the *test set*, including the remaining 150 individuals. The development set is used to tune the parameters of the verification system and of the fusion experiments done in this paper (indicated later in this Section). No training of parameters is done on the test set. The following matchings are defined in each set: *a*) genuine matchings: the 4 samples in the first session to the 4 samples in the second session; and *b*) impostor matchings: the 4 samples in the first session to 1 sample in the second session of the remaining users. With the development set, this results in 50 individuals $\times$ 2 eyes $\times$ 4 templates $\times$ 4 test images $= 1,600$ genuine scores, and 50 individuals $\times$ 2 eyes $\times$ 4 templates $\times$ 49 test images $= 19,600$ impostor scores. Similarly, for the test set we have 150 individuals $\times$ 2 eyes $\times$ 4 templates $\times$ 4 test images $= 4,800$ genuine scores, and 150 individuals $\times$ 2 eyes $\times$ 4 templates $\times$ 149 test images $= 178,800$ impostor scores.

We have automatically segmented all the iris images using circular Hough transform in order to detect the iris and pupil boundaries, which are modeled as two concentric circles [4]. Then, automatically segmented images have been visually inspected to manually correct images not well segmented. With this procedure, we obtain a correct segmentation of the 100% of the database. The objective is avoid bias in the matching performance due to incorrectly segmented images.

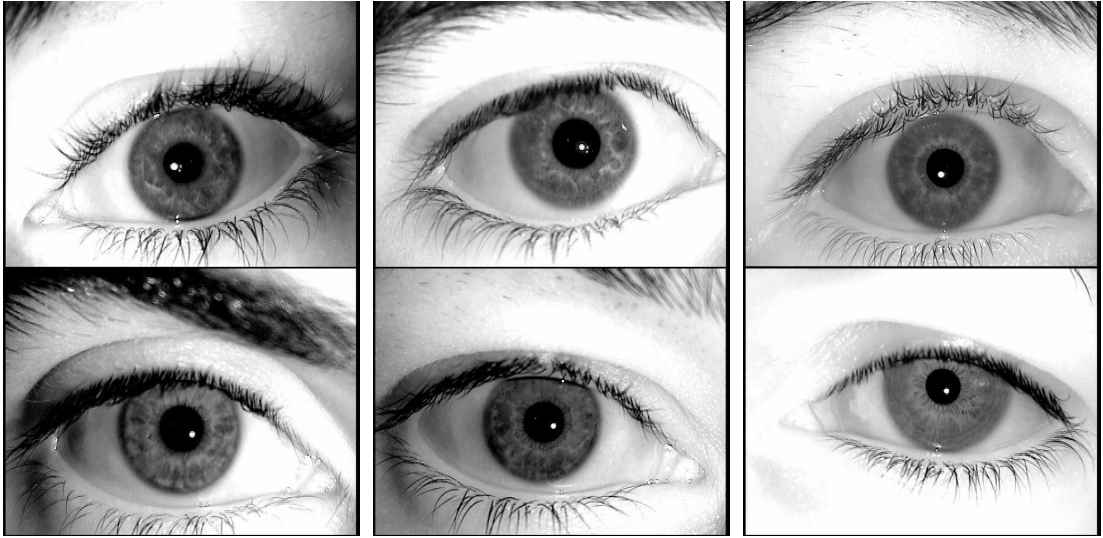

Fig. 6. Iris examples from the BioSec database.

We then construct a binary mask that includes only the iris region and use it to discard SIFT keypoints being detected outside the mask. An example of segmented images together with the detected SIFT keypoints can be seen in Figure 7. Since eyelash and eyelid occlusion is not very prominent in our database, no technique was implemented to detect eyelashes or eyelids.

*B. Baseline iris matcher*

This system performs a normalization of the segmented iris region by using a technique based on Daugman's rubber sheet model [4]. The centre of the pupil is considered as the reference point, and radial vectors pass through the iris region. Since the pupil can be non-concentric to the iris, a remapping formula for rescale points depending on the angle around the circle is used. Normalization produces a 2D array with horizontal dimensions of angular resolution and vertical dimensions of radial resolution. This normalization step is as shown in Figure 1.

Feature encoding is implemented by convolving the normalized iris pattern with 1D Log-Gabor wavelets. The 2D normalized pattern is broken up into a number of 1D signals, and then these 1D signals are convolved with 1D Gabor wavelets. The rows of the 2D normalized pattern are taken as the 1D signal, each row corresponds to a circular ring on the iris region. It uses the angular direction since maximum independence occurs in this direction [12].

[1] The source code can be freely downloaded from www.csse.uwa.edu.au/~pk/studentprojects/libor/sourcecode.html

The output of filtering is then phase quantized to four levels using the Daugman method [4], with each filtering producing two bits of data. The output of phase quantization is a grey code, so that when going from one quadrant to another, only 1 bit changes. This will minimize the number of bits disagreeing, if say two intra-class patterns are slightly misaligned and thus will provide more accurate recognition [12]. The encoding process produces a bitwise template containing a number of bits of information.

For matching, the Hamming distance (HD) is chosen as a metric for recognition, since bitwise comparisons are necessary. In order to account for rotational inconsistencies, when the Hamming distance of two templates is calculated, one template is shifted left and right bitwise and a number of Hamming distance values is calculated from successive shifts [4]. This method corrects for misalignments in the normalized iris pattern caused by rotational differences during imaging. From the calculated distance values, the lowest one is taken.

*C. Results*

First, the SIFT matcher is optimized in terms of its different parameters. The experimental parameters to be set are: the scale factor of the Gaussian function $\sigma=1.6$; the number of scales $s=3$; the threshold $D$ excluding low contrast points; the threshold $r$ excluding edge points ($r=10$); the threshold of the ratio $d_1/d_2$ (set to $0.76$) and the tolerances $\varepsilon_\theta$ and $\varepsilon$ for trimming of false matches. The indicated values of the parameters have been extracted from Lowe [7]. We have noted however that the threshold $D$ indicated in [7] discards too many SIFT keypoints of the iris region ($D=0.03$ when pixel values are in the range [0,1]). Thus, together with $\varepsilon_\theta$ and $\varepsilon$, we have decided to find an optimal value also for $D$.

Figure 8 shows the verification performance of our SIFT implementation on the development set in terms of EER (%) as we vary $\varepsilon_\theta$ and $\varepsilon$ when $D=0.25/255$, $D=0.5/255$ and $D$=0.75/255. The optimal combination of parameters in these three cases (i.e. those that results in the lowest EER) are also summarized in Table I, together with the case where no trimming of false matches is carried out. We observe that by trimming out false matches using geometric constraints, the EER is reduced to the fourth part.

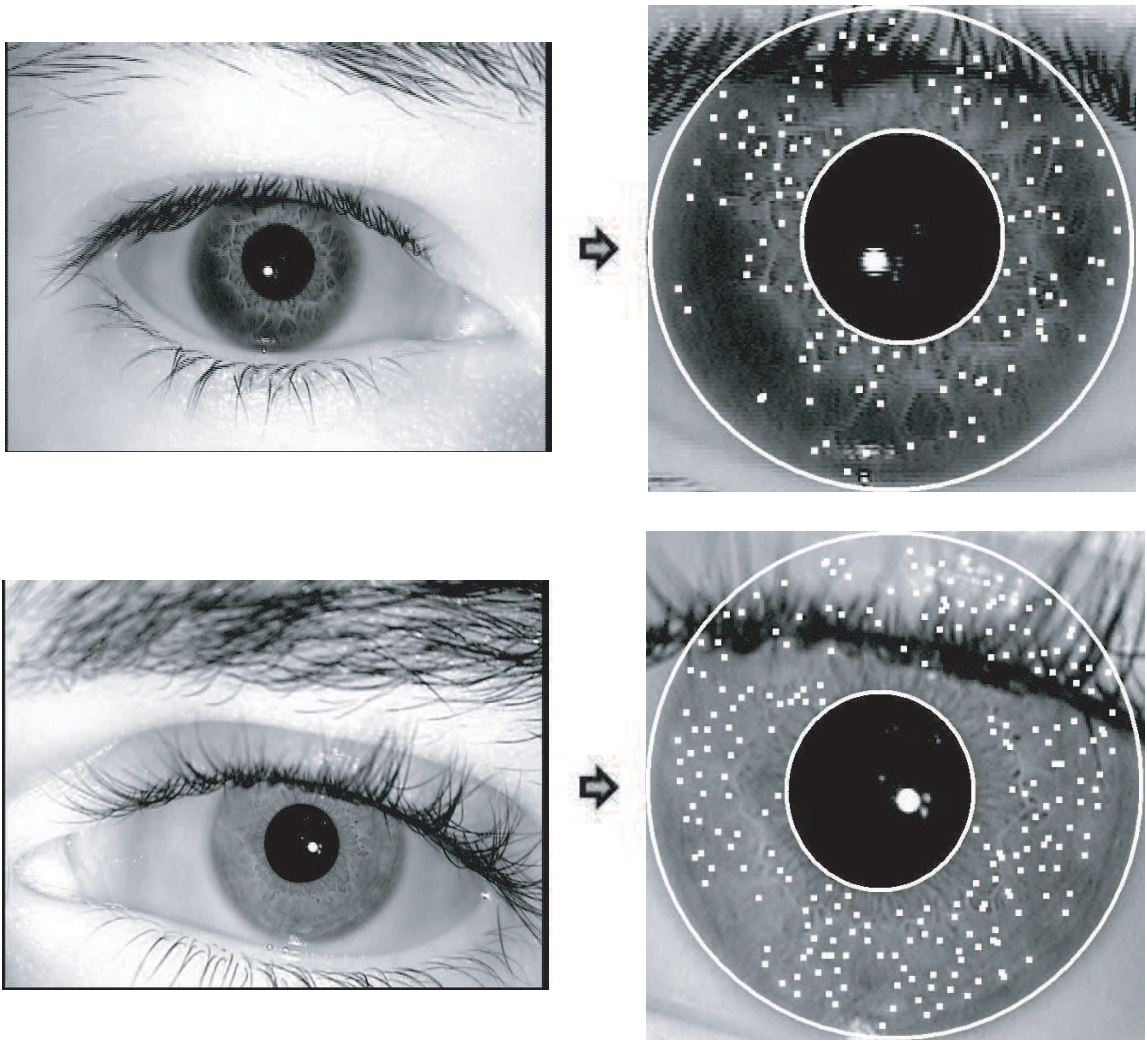

Fig. 7. Example of segmented images together with their detected SIFT keypoints.

Based on the results of Figure 8 and Table I, the best combination of parameters is therefore $D$=0.25/255, $\varepsilon_\theta$=18 and $\varepsilon$ =14. Figure 9 depicts the performance of the SIFT matcher for this case. We observe that the optimal value of $D$ in our SIFT implementation, $D$=0.25/255=0.00098, is much lower than 0.03 (as recommended in [7]). Concerning the values of the tolerances $\varepsilon_\theta$ and $\varepsilon$, it can be seen in Figure 8 that the EER monotonically decreases as the two tolerances are increased until a minimum in the EER is reached (the exact values of $\varepsilon_\theta$ and $\varepsilon$ at the minimum are indicated in Table I). Once this minimum is reached, the EER is slightly increased again with the tolerances.

We now compare the performance of our SIFT implementation with the baseline iris matcher of Section III-B. Figure 10 comparatively shows the performance of the two matchers using DET curves, both on the development and on the test set. We also have performed a fusion of the SIFT and baseline matchers using sum rule with tanh normalization [15]:

$$s^t = \frac{1}{2}\left(\tanh\left(0.01\frac{s - \mu_s}{\sigma}\right) + 1\right) \quad (3)$$

where $s$ is the raw similarity score, $s^t$ denotes the normalized similarity score, and $\mu_s$ and $\sigma_s$ are respectively the estimated mean and standard deviation of the genuine score distribution. Table II summarizes the Equal Error Rates (EER) computed from Figure 10. We observe that the fusion of the two matchers results in better performance than either of the two matchers,

IV. CONCLUSIONS AND FUTURE WORK

In this paper, we have proposed the use of the SIFT operator for iris feature extraction and matching. There have been a few studies using SIFT for face [8], [9] and fingerprint [10] recognition, and some recent studies also for iris [5]. In this work, we contribute with the analysis of the influence of

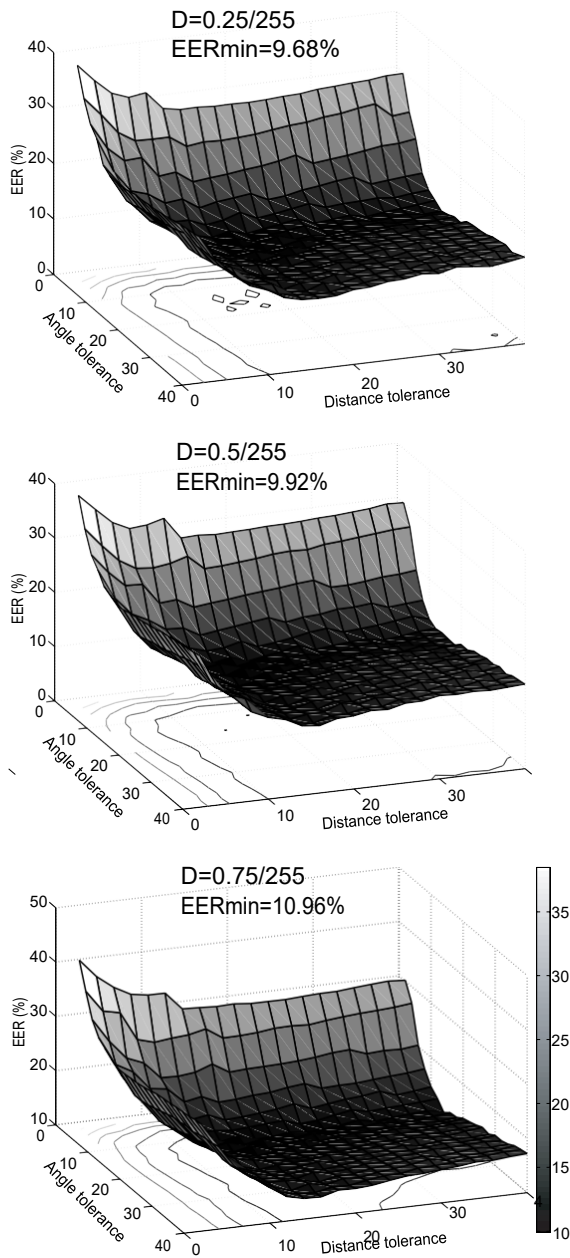

Fig. 8. Development set. Verification results of the SIFT matcher in terms of EER (%) depending on the threshold $D$ and the tolerances of angle ($\varepsilon_\theta$) and distance ($\varepsilon$).

| $D$ | $\varepsilon_\theta$ | $\varepsilon$ | EER |
|---|---|---|---|
| 0.25 | - | - | 36.85% |
| **0.25** | **18** | **14** | **9.68%** |
| 0.5 | 14 | 16 | 9.92% |
| 0.75 | 18 | 14 | 10.96% |
| 1 | 16 | 14 | 14.03% |

TABLE I

DEVELOPMENT SET - SIFT MATCHER. OPTIMAL COMBINATIONS OF THE PARAMETERS $D$ AND TOLERANCES OF ANGLE ($\varepsilon_\theta$) AND DISTANCE ($\varepsilon$). THE COMBINATION RESULTING IN THE LOWEST EER IS MARKED IN BOLD. THE FIRST ROW INDICATES THE CASE WHERE NO TRIMMING OF FALSE MATCHES IS CARRIED OUT.

|  | SIFT | Baseline | Fusion |
|---|---|---|---|
| Development set | 9.68% | 4.64% | - |
| Test set | 11.52% | 3.89% | 2.96% |

TABLE II

EER OF SIFT, BASELINE AND FUSION MATCHERS.

different SIFT parameters on the verification performance, including trimming of false matches with geometric constraints, as proposed in [10] for the case of fingerprints. Although the performance of our implementation is below popular matching approaches based on transformation to polar coordinates and Log-Gabor wavelets, we also show that their fusion provides a performance improvement of 24% in the EER. This is because the sources of information used in the two matchers are different, providing complementary sources of information.

Future work will be focused on the improvement of the SIFT matcher by detecting eyelids, eyelashes and specular reflections [6], thus discarding SIFT keypoints computed in these regions. We are also working on the inclusion of local iris quality measures [16] to account for the reliability of extracted SIFT points, so if the quality is high for two matched points, they will contribute more to the computation of the matching score.

Current iris recognition systems based on accurate segmentation and transformation to polar coordinates rely on cooperative data, where the irises have centered gaze, little eyelashes or eyelids occlusion, and illumination is fairly constant [5]. The SIFT-based method does not require polar transformation or highly accurate segmentation, and it is invariant to illumination, scale, rotation and affine transformations [7]. This makes the SIFT approach feasible for biometric recognition of distant and moving people, e.g. the "Iris on the Move" project [14], where a person is recognized while walking at normal speed through an access control point such as those common at airports. Currently this is one of the research hottest topics within the international biometric community [17], which drastically reduces the need of user's cooperation, and it will be another important source of future work.

## V. ACKNOWLEDGMENTS

This work has been supported by Spanish MCYT TEC2006-13141-C03-03 project. Author F. A.-F. is supported by a Juan de la Cierva Fellowship from the Spanish MICINN.

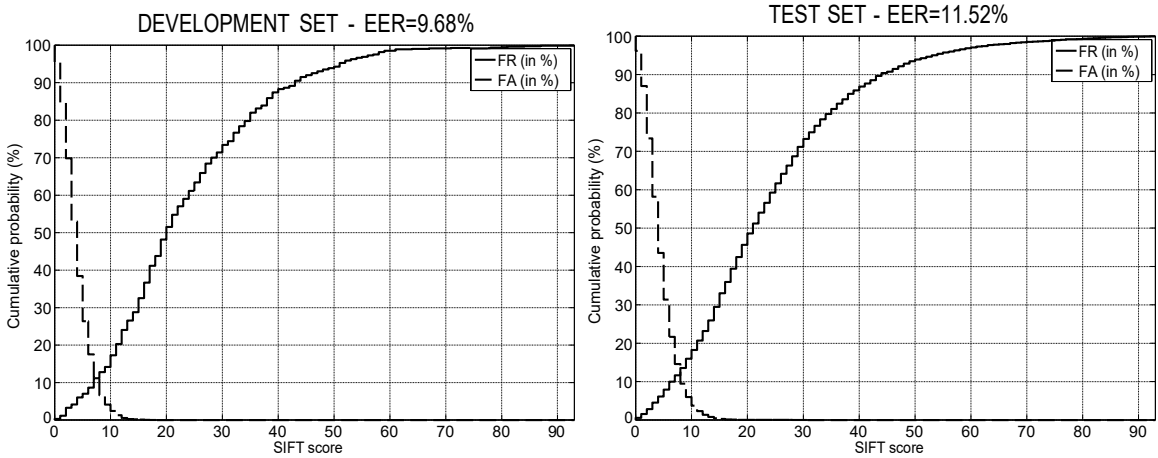

Fig. 9. Performance of the SIFT matcher (FR=False Rejection, FA=False Acceptance).

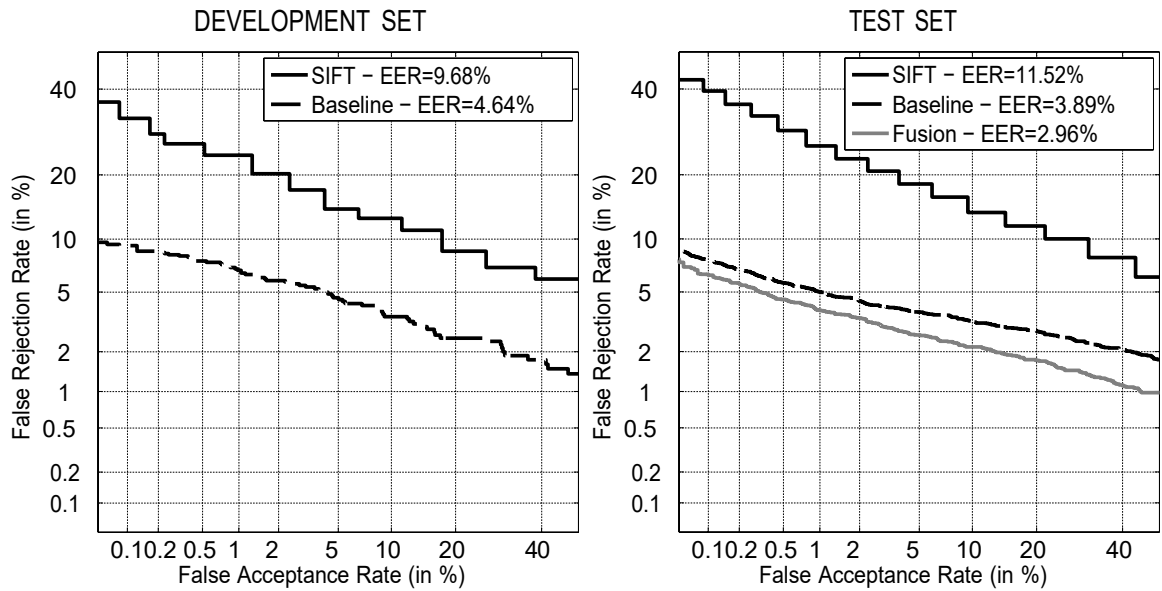

Fig. 10. Performance of the SIFT and the baseline matchers and their fusion results.